\documentclass[journal]{IEEEtran}
\IEEEoverridecommandlockouts

\usepackage{cite}
\usepackage{amsmath,amssymb,amsfonts}
\usepackage{algorithmic}
\usepackage{graphicx}
\usepackage{textcomp}
\usepackage{xcolor}
\def\BibTeX{{\rm B\kern-.05em{\sc i\kern-.025em b}\kern-.08em
    T\kern-.1667em\lower.7ex\hbox{E}\kern-.125emX}}

\usepackage{epstopdf}
\usepackage{lipsum}
\usepackage{makecell}
\usepackage{pdfpages}
\usepackage{tikz}
\usetikzlibrary{shapes,arrows}
\usepackage{pgf-pie}
\usepackage{pgfplots}
\usepgfplotslibrary{groupplots}
\usepackage{verbatim}
\pgfplotsset{compat=1.17}
\usepackage{threeparttable}
\usepackage{helvet}
\usepackage[eulergreek]{sansmath}

\usepackage{tabularx} 
\usepackage{multirow} 
\usepackage{subfig}

\usepackage{siunitx}

\usepackage{pifont}
\usepackage{flushend}

\usepackage{multirow}

\usepackage{hyperref}
\usepackage{textpos}

\begin{document}
    \bstctlcite{IEEEexample:BSTcontrol}

	\title{
	Towards Hardware Supported Domain Generalization in DNN-based Edge Computing Devices for Health Monitoring\\

	\thanks{The authors are with the Chair of Integrated Digital Systems and Circuit Design, RWTH Aachen University, 52074 Aachen, Germany (e-mail: loh@ids.rwth-aachen.de; gemmeke@ids.rwth-aachen.de).}
	\thanks{This work is partially funded by the German Federal Ministry of Education and Research under grant no. 03ZU1106CA (Clusters4Future – NeuroSys) and partially by the German Federal Ministry for the Environment, Nature Conservation, Nuclear Safety and Consumer Protection under grant no. 67KI32006A (RESCALE).}
	}

	\author{
 		Johnson Loh, Lyubov Dudchenko, Justus Viga and Tobias Gemmeke, \textit{Senior Member, IEEE}
	}
	
	\maketitle

\begin{textblock*}{\textwidth}(-0.5cm,-7cm) 
\small{© 2024 IEEE. Personal use of this material is permitted. Permission from IEEE must be obtained for all other uses, in any current or future media, including reprinting/republishing this material for advertising or promotional purposes, creating new collective works, for resale or redistribution to servers or lists, or reuse of any copyrighted component of this work in other works. This paper was accepted in IEEE Transactions on Biomedical Circuits and Systems. DOI: \href{https://doi.org/10.1109/TBCAS.2024.3418085}{10.1109/TBCAS.2024.3418085}} \vspace{0.1cm} 
\end{textblock*}

\begin{abstract}
    Deep neural network (DNN) models have shown remarkable success in many real-world scenarios, such as object detection and classification.
    Unfortunately, these models are not yet widely adopted in health monitoring due to exceptionally high requirements for model robustness and deployment in highly resource-constrained devices. 
    In particular, the acquisition of biosignals, such as electrocardiogram (ECG), is subject to large variations between training and deployment, necessitating domain generalization (DG) for robust classification quality across sensors and patients. 
    The continuous monitoring of ECG also requires the execution of DNN models in convenient wearable devices, which is achieved by specialized ECG accelerators with small form factor and ultra-low power consumption. 
    However, combining DG capabilities with ECG accelerators remains a challenge. 
    This article provides a comprehensive overview of ECG accelerators and DG methods and discusses the implication of the combination of both domains, such that multi-domain ECG monitoring is enabled with emerging algorithm-hardware co-optimized systems.
    Within this context, an approach based on correction layers is proposed to deploy DG capabilities on the edge.
    Here, the DNN fine-tuning for unknown domains is limited to a single layer, while the remaining DNN model remains unmodified.
    Thus, computational complexity (CC) for DG is reduced with minimal memory overhead compared to conventional fine-tuning of the whole DNN model.
    The DNN model-dependent CC is reduced by more than {2.5\,\texttimes} compared to DNN fine-tuning at an average increase of F1 score by more than {20\,\%} on the generalized target domain.
    In summary, this article provides a novel perspective on robust DNN classification on the edge for health monitoring applications.
\end{abstract}



\section{Introduction} 
\label{sec:intro}

\IEEEPARstart{A}{s} wearable devices increasingly become available in people's daily lives, there is a growing need for edge computing devices capable of analysing personal data in a privacy preserving system.
Complex machine learning models fitted to the edge enable wearable devices with advanced processing capabilities without transmission of personal data to third-party systems. 
Deep neural networks (DNNs) have shown impressive results in several applications, such as image classification \cite{Krizhevsky2017} and speech recognition \cite{Zeineldeen2022}.
Nevertheless, their high computational complexity (CC) is still a challenge for deployment in resource-constrained edge devices. 
To address this challenge, DNN accelerators have been developed to enable the efficient execution of large DNN models on specialized hardware \cite{Latotzke2021}.
Although current DNN accelerators can perform inference tasks efficiently, in practice, the features in the actual data deviate from those in the data used during training - the \textit{domain shift} problem.
In practical settings, the domain shift is caused by differences in the background environment, measurement setup, or differences in the measured subject.



\begin{figure*}[t]
    \centering
    \includegraphics[width=\textwidth]{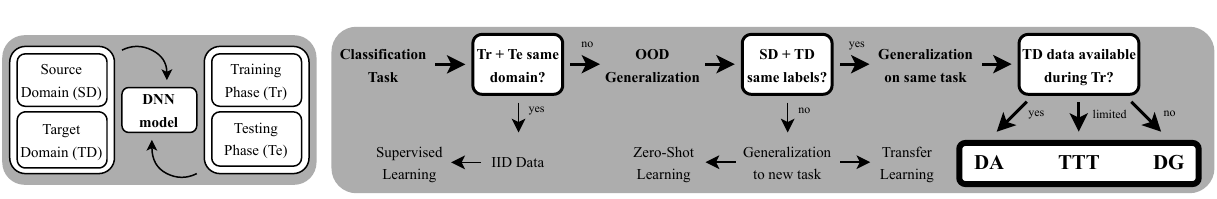}
    \caption{Problem Overview}
    \label{fig:problem}
\end{figure*}

The generalization to these out-of-distribution (OOD) data is well addressed for a wide range of applications \cite{Quinonero-Candela2022,Zhou2022}, in particular image processing tasks \cite{Yosinski2014,Zhao2022}.
Figure \ref{fig:problem} shows a brief summary of the problem of OOD generalization.
While classical DNN training assumes independent and identically distributed (IID) data, i.\,e. source domain (SD) and target domain (TD) are similarly distributed, the OOD generalization problem introduces a discrepancy between SD and TD.
Typically, data from SD are utilized to train the DNN during a training phase, while data from TD is used to evaluate the quality of generalization during a testing phase.
Dependent on the availability of labels in SD/TD and data during training, the general field of OOD generalization can be further subdivided into e.g. detection of new classes (Zero-Shot Learning \cite{Pourpanah2022}) or transfer of features to new classification tasks (Transfer Learning \cite{Zhuang2021}).
Here, domain generalization (DG) deals with the increased robustness against domain shift, while the classification task remains the same as during training.
Especially, working with faint features as in electrocardiogram (ECG) signals requires generalizing properties as domain shift heavily affects the performance of pre-trained DNN models \cite{Seo2021}.


In this case the challenge is the support of DG in wearables, which go beyond DNN parameter reconfiguration or on-line DNN training as supported in state-of-the-art ECG accelerators.
To the best of our knowledge, there is no comprehensive work addressing the acceleration of DG-based algorithms in hardware (HW) and, hence, this field poses a new and promising direction for future DNN systems.
The application case of ECG monitoring is well suited for an initial feasibility study, as large domain variations exist between collected databases and actual deployment on wearables.
Further, on-device processing is necessary to ensure that personal data is handled locally.


Therefore, this work summarizes DG methods and analyses them with the intention for deployment on dedicated accelerators in edge devices.
First, the state-of-the-art is reviewed to compile common features and trends for DG in general and specific to ECG signals.
Then, the deployment scenarios are outlined to investigate the impact on HW resources and communication (see Section\,\ref{sec:background}).
As an initial entry point for DG specific accelerators, the concept of correction layers (CLs) are introduced as a means for DG with minimal HW overhead (see Section\,\ref{sec:cl_alg}).
For atrial fibrillation classification, we show statistically robust classification performance improvements, while the original model is unmodified and augmented with a single layer.
The evaluation of memory and computational complexity shows that CL outperforms conventional DNN fine-tuning, while quantitative gains are dependent on DNN architecture and CL position (Section\,\ref{sec:cl_hw}).
This minimally invasive approach serves as a baseline for further improvements on DG implementations through algorithm-hardware co-optimizations.

The main contributions of this work are summarized as follows:
\begin{itemize}
	\item Review of DG methods is provided in the context of ECG classification for deployment on dedicated HW accelerators 
	\item Correction layers (CLs) are proposed in a case study to perform DG with minimal modifications on pre-trained models and robust classification performance 
	\item Evaluation of CL performance during inference using implemented HW accelerators synthesized in 22\,nm CMOS
	\item Estimation of memory and computational complexity of CL during training 
\end{itemize}

\section{Background} 
\label{sec:background}


As indicated in Section \ref{sec:intro}, the spectrum of DG methods is large, while only a subset is currently applied on ECG classification in state-of-the-art literature.
Hence, this section briefly introduces existing state-of-the-art methods and analyses constraints in their applicability to the ECG domain and for HW acceleration.

\subsection{DG method overview}

As depicted in Fig. \ref{fig:problem}, OOD generalization covers multiple application fields, while the generalization for the sake of model robustness, i.e. classification of same classes in SD and TD, also differentiates between domain adaptation (DA), test-time training (TTT) and DG.
The main difference is the availability of TD data during the training process.
The extreme cases assume the corner scenarios, where full \cite{Long2015} or no information \cite{Muandet2013} in the TD is available, respectively.
Adaptations during test-time, however, uses limited data from TD to fine-tune the model during deployment \cite{Sun2020}.
As the target of this work is to increase ECG classification robustness, we look at both corner cases and refer to both as DG for the sake of simplicity.


\newcommand{\PreserveBackslash}[1]{\let\temp=\\#1\let\\=\temp}
\newcolumntype{C}[1]{>{\PreserveBackslash\centering}m{#1}}
\begin{table}[tb]
	\centering
	\begin{threeparttable}[b]
	\caption{Required component modifications during DNN training for DG algorithms}
	\label{tab:baseline_comparison}
	\begin{tabular}{C{0.1\textwidth}||C{0.033\textwidth}|C{0.033\textwidth}|C{0.033\textwidth}|C{0.033\textwidth}|C{0.033\textwidth}|C{0.033\textwidth}}
		 & Input data & Labels & Addtl. Models & DNN Arch. & DNN param. & optim. function \\
		\hline
		\hline
		Domain Alignment & \ding{55} & \ding{55} & \ding{55} & N/A & \ding{51} & \ding{51} \\
		\hline
		Meta-Learning & N/A & N/A & \ding{51} & N/A & \ding{51} & \ding{51} \\
		\hline
		Ensemble Learning & \ding{55} & \ding{55} & \ding{55} & \ding{51} & \ding{51} & N/A \\
		\hline
		Data Augmentation & \ding{51} & \ding{51} & \ding{55} & \ding{55} & \ding{51} & \ding{55} \\
		\hline
		Self-Supervised Learning & \ding{51} & \ding{51} & \ding{55} & \ding{55} & \ding{51} & N/A \\
		\hline
		Disentangled Representation & \ding{55} & \ding{55} & \ding{55} & N/A & \ding{51} & \ding{51} \\
		\hline
		Regularization & \ding{55} & \ding{55} & \ding{55} & \ding{55} & \ding{51} & \ding{51} \\
		\hline
		Expert knowledge & \ding{55} & \ding{55} & \ding{51} & N/A & \ding{55} & \ding{55} \\
	\end{tabular}
	\end{threeparttable}
\end{table}

\begin{figure*}[t]
    \centering
    \includegraphics[width=\textwidth]{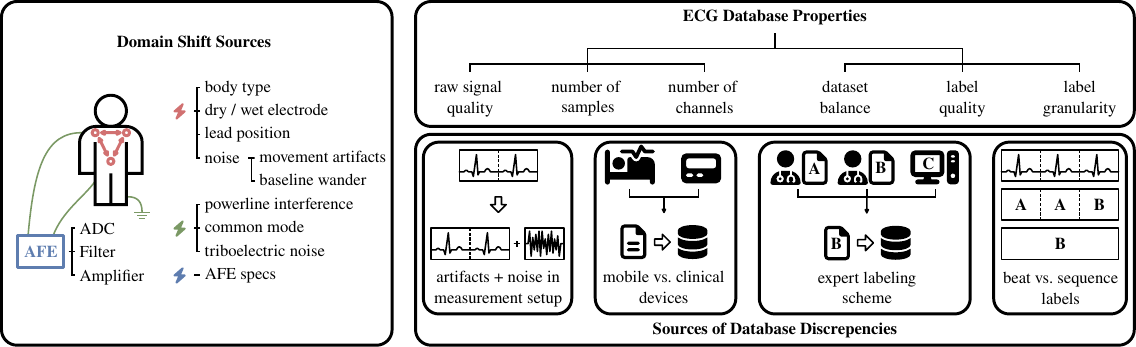}
    \caption{An overview on the sources of domain shift in the application of ECG monitoring and the challenges of ECG data acquisition in general. Many databases differ in both measurement setup and labeling schemes making it challenging to combine them for DG.}
    \label{fig:ds_overview}
\end{figure*}

Table \ref{tab:baseline_comparison} summarizes recent DG methods.
Starting from a trained DNN model, modifications on the reference are performed.
Some approaches, such as data augmentation and self-supervised learning, aim to extend the training dataset with additional synthetic \cite{Shorten2019} or unlabeled samples \cite{Krishnan2022}, which enclose the features of TD.
Other approaches, such as meta-learning, require additional models to incorporate domain shift in the training process \cite{Li2018}.
Modifications on the original DNN model architecture are also utilized, e.g. with ensembles of domain specific classifiers \cite{Deng2021}.
Another common method is to modify the training algorithm, e.g. by including additional loss terms to achieve domain-invariant features \cite{Bazi2013,Chen2020,Wang2021}.
In general, most methods require an update of DNN model parameters to adjust for the new TD.
In the perspective of DG HW acceleration, methods will be preferred, which require least modification on the original inference model to calculate the DG algorithm.
Here, DG complexity mainly depends on the target application. 


\subsection{DG applied on ECG}
\label{ssec:ecg_dg}

In this work, we investigate ECG classification as an example application.
Figure \ref{fig:ds_overview} summarizes this application field in terms of available data and what sources of noise and domain shift is captured in the data.
An important point to consider in the complexity of DG is that domain shift is not only caused by inter-patient differences \cite{Sraitih2021}, but also the experimental setup, which ranges from variations in lead positioning to contact impedance of electrodes resulting in deviating system responses for the captured data.
The analog front end (AFE) further influences signal quality and introduces domain shift, when AFE configurations, i.e. ADC resolution etc., differ.

In principle, these perturbations can be modeled in a large enough corpus of datasets, which incorporates all varieties with sufficient sample sizes.
However, biomedical data suffer from small number of labeled data \cite{Mehari2022} and  unbalanced distribution of classes \cite{Wang2021b}.
This is partially resulting from the scarcity of anomalies and the complexity of defining features, which even trained experts struggle to identify reliably \cite{Kirchhof2016}.
The annotation of large amounts of data require large resources in terms of trained experts and time, which potentially affects the quality of provided labels \cite{Clifford2017}.
Discrepancies in the labeling paradigm, e.g. beat vs. sequence labels or hybrid expert and machine-based labeling further complicate the combination of multiple datasets for training.
Therefore, DG algorithm development mainly focus on homogeneous domain shift within one dataset, i.e. inter-patient paradigm \cite{Sraitih2021}, or multiple datasets with similiar quality, i.e. (near) clinical measurement setup \cite{Alday2020}.

\begin{figure}[ht]
    \centering
  \subfloat[Extension of an active dataset for model re-training\label{fig:ecg_sota_a}]{%
      \includegraphics[width=0.45\linewidth]{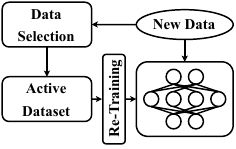}}
    \hfill
  \subfloat[Domain alignment and separation in feature space\label{fig:ecg_sota_b}]{%
        \includegraphics[width=0.45\linewidth]{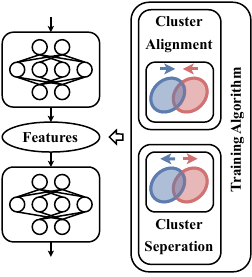}}
    \\
  \subfloat[Ensemble classifiers with domain invariant and domain specific subbranches\label{fig:ecg_sota_c}]{%
        \includegraphics[width=0.7\linewidth]{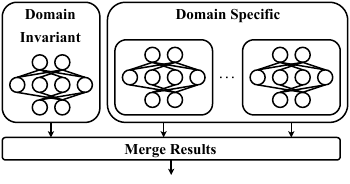}}
  \caption{Sketch of state-of-the-art DG algorithms for ECG classification}
  \label{fig:ecg_sota} 
\end{figure}

Figure \ref{fig:ecg_sota} shows a conceptual overview of DG algorithms applied in ECG classification.
One method is to select relevant data samples and add them into an active dataset, which is used to fine-tune or re-train the DNN \cite{Rahhal2016,Wang2019}.
Since the new data incorporate samples in the target domain, the DNN is continuously adjusted towards new data.
Another method adjusts the training process, such that features from different domains are aligned to each other, while features from different classes are separated \cite{Bazi2013,Ganin2016,Yin2019,Chen2020,Hasani2020,Shang2021,Wang2021}.
The main idea is that feature distributions from SD and TD are similar and should be robust against domain shift.
The alignment process creates domain invariant features by minimizing distance metrics of those feature distributions.
A third concept employs ensembles of classifiers, which classify each domain independently from each other through domain-specific branches in addition to a general domain-invariant branch \cite{Deng2021}.


\subsection{ECG accelerators}

Classical ECG accelerators mainly focused on the efficient execution of classification models.
These models range from traditional thresholding of ECG features, such as RR intervals \cite{Chen2015}, to recently exotic classifiers, such as spiking neural networks \cite{Bauer2019,Chu2022,Liu2022,Mao2022}.
However, the majority of works investigates DNN variants such as convolutional neural networks (CNNs) \cite{Liu2021,Wang2021a,Loh2022,Lu2022}, multi-layer perceptrons (MLPs) \cite{Yin2019a,Zhao2020,Parmar2023}, long-short term memories (LSTMs) \cite{Sivapalan2022} and gated recurrent units (GRUs) \cite{Jobst2022}.
While all works aim to achieve ultra-low power consumption or energy per solution, the adopted hardware platform, i.e. ARM processor \cite{Sivapalan2022}, FPGA \cite{Lu2022} and ASICs \cite{Yin2019a,Zhao2020,Liu2021,Wang2021a,Jobst2022,Loh2022}, varies.
The complexity of the algorithm and their system architecture play a critical role in the choice.
While processor architectures and FPGAs offer high flexibility in terms of programmability, the power consumption is generally several orders of magnitude higher than ASIC implementations ($>$\SI{}{\micro\watt} \cite{Lu2022,Sivapalan2022} vs. \SI{}{\nano\watt} range \cite{Jobst2022,Loh2022}).
Nevertheless, ASIC inference engines offer certain reconfigurability for model weights and architecture \cite{Liu2021,Jobst2022,Loh2022}.
However, the architecture reconfiguration is usually limited to layer depth and width of convolution-based DNN layers.
Generally, other processing components, such feature extraction, are fixed by the implemented architecture.



	\begin{figure*}
    \centering
    \includegraphics[width = 0.95\textwidth]{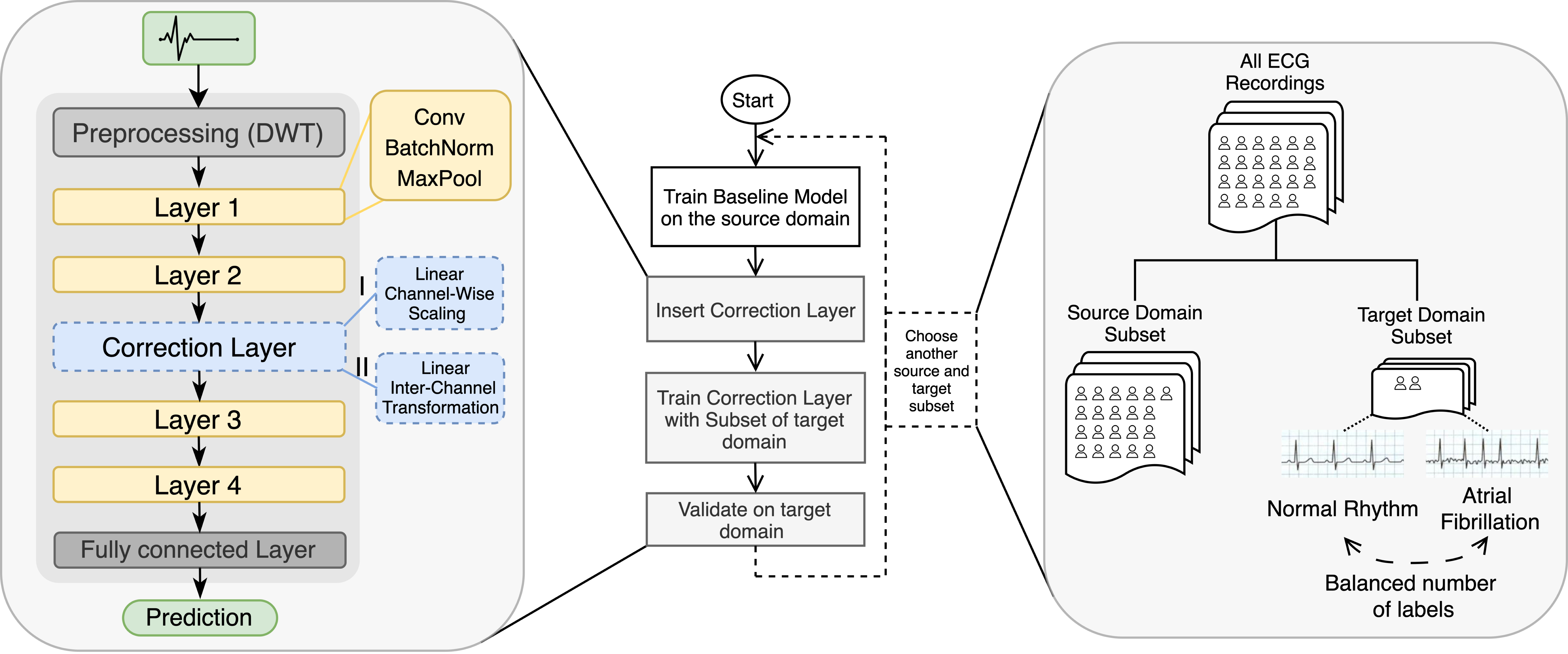}
    \caption{Insertion of Correction Layer into CNN from \cite{Loh2022}. The training and validation of DG capability is performed in two stages. The first stage trains the CNN without CL on SD. The second stage trains the CL only on a subset of TD, while the remainder of TD is used for validation. Both stages are performed for different combinations of patients in SD and are cross-validated for statistically robust QoS evaluation.}
    \label{fig:cl_concept}
\end{figure*}

\section{Correction Layer Approach} 
\label{sec:cl_alg}

In this section, we detail the concept of CL as one additionally inserted layer in a pre-trained DNN.
Here, the free trainable parameters of the CL are used to perform DG, while the pre-trained DNN remains frozen.
As the training is limited to the CL only, we expect significant CC reduction compared to fine-tuning the whole DNN.

\subsection{Motivation}
\label{ssec:da_compl}

Considering previous DG works (as summarized in Section \ref{ssec:ecg_dg}), it is evident that computational resources are not the primary optimization objective, since the extension of training dataset with new samples and domain specific DNN branches require scalable memory and computation resources.
Furthermore, fine-tuning a DNN using domain alignment techniques on the edge either require on-device learning \cite{Lee2021} or a distributed training setup to update all DNN parameters.
Within this work, we investigate DG from the perspective of constrained complexity.
A critical question is how much complexity is necessary to achieve generalization across domains.

\begin{figure}[ht]
    \centering
    \includegraphics[width=0.485\textwidth]{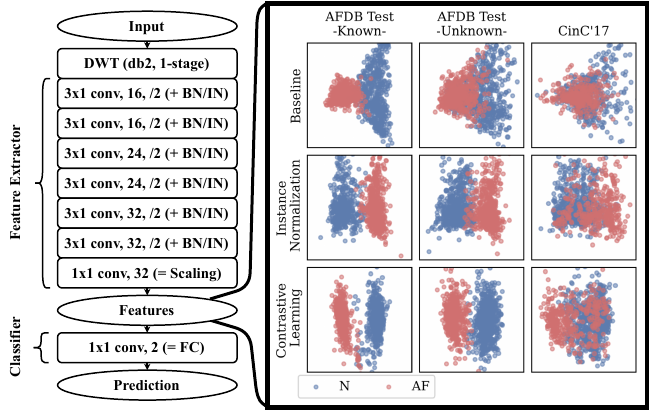}
    \caption{Feature distribution of a fully convolutional DNN for IN and CTL. Features are mapped to the 2D plane using principle component analysis (PCA).}
    \label{fig:feat_distr}
\end{figure}

To answer this question, we performed an experiment to evaluate the necessary complexity to achieve generalization for homogeneous and heterogeneous domain shift.
In specific, we focused on the binary classification of atrial fibrillation (AF) and other (mainly normal) signals using a fully convolutional DNN.
Training and validation is performed on the MIT-BIH Atrial Fibrillation Database (AFDB) \cite{Moody1983} using 5-fold crossvalidation, while the Computing in Cardiology Challenge 2017 (CinC'17) \cite{Clifford2017} is added in the validation set.
Here, inter-patient deviations in the dataset samples is used to incorporate homogeneous domain shift, where each fold contains a disjunct set of patients to guarantee unseen patients in the validation set.
Due to the different data acquisition setup of the CinC'17 benchmark, those samples represent the heterogeneous domain shift compared to the AFDB reference.
Figure \ref{fig:feat_distr} shows the employed network architecture and the feature distribution for the different datasets and the different DG techniques.
Specifically, we chose patient-specific instance normalization (IN) as a regularization method with minimal additional parameters and contrastive learning (CTL) for feature alignment/separation using contrastive losses \cite{Hadsell2006}.

Table \ref{tab:dg_quality_domain_shift} shows the detailed quality of service (QoS) of the classification.
It is evident, that the baseline without DG is less accurate on unseen samples and performs poorly on CinC'17.
IN and CTL improved QoS significantly in CinC'17, especially for AF.
The feature distribution also indicates that a clear clustering of features is achieved through both methods.
However, the QoS difference between IN and CTL is small, which motivates an emerging class of algorithms for HW co-optimization to fill the gap between high quality DG and no model finetuning.
Inspired by the simplicity of IN as a linear transformation in each layer, we introduce correction layer (CL) as a single layer alternative in Section \ref{ssec:cl_alg_method}.
The key principle is to limit the modification to the addition of a single CL, such that the original DNN architecture and parameters remain untouched.
This enables the reduction of CC and memory for training, which is reduced from the whole network to a single layer.

\begin{table}[tb]
	\centering
	\begin{threeparttable}[b]
	\caption{Validation F1 score of DG techniques based on homogeneous (AFDB) and heterogeneous (CinC'17) domain shift using 5-fold cross-validation.}
	\label{tab:dg_quality_domain_shift}
	\begin{tabularx}{0.485\textwidth}{C{0.07\textwidth}||C{0.036\textwidth}|C{0.036\textwidth}|C{0.036\textwidth}|C{0.036\textwidth}|C{0.036\textwidth}|C{0.036\textwidth}}
         & \multicolumn{4}{c|}{AFDB} & \multicolumn{2}{c}{CinC'17} \\
         \cline{2-5}\cline{5-7}
         & \multicolumn{2}{c|}{Test - Known} & \multicolumn{2}{c|}{Test - Unknown} & \multicolumn{2}{c}{Test - Unknown} \\
		 & N & AF & N & AF & N & AF \\
		\hline
		\hline
		Baseline (w/o DG) & 0.93 \textpm\,0.03 & 0.90 \textpm\,0.03 & 0.91 \textpm\,0.07 & 0.88 \textpm\,0.07 & 0.67 \textpm\,0.01 & 0.17 \textpm\,0.21 \\
		\hline
		Instance Norm. & 0.96 \textpm\,0.02 & 0.93 \textpm\,0.03 & 0.92 \textpm\,0.10 & 0.91 \textpm\,0.09 & 0.79 \textpm\,0.02 & 0.78 \textpm\,0.03 \\
		\hline
		Contrastive Learning & 0.97 \textpm\,0.02 & 0.94 \textpm\,0.02 & 0.96 \textpm\,0.04 & 0.94 \textpm\,0.05 & 0.80 \textpm\,0.01 & 0.77 \textpm\,0.02 \\

	\end{tabularx}
	\end{threeparttable}
\end{table}

\subsection{CL Algorithm}
\label{ssec:cl_alg_method}

The general idea of CL is to insert a single layer into an existing DNN to transform the intermediate activations, such that the features after transformation are robust against domain shift.
The layers themselves contain trainable parameters, which can be adjusted both in a supervised and unsupervised manner.
Note that the resulting robust features can follow domain-specific models, such as language models in automatic speech recognition \cite{Li2022}, but some applications, e.g. ECG classification, such domain-invariant expert features are not as clearly defined. 
From the structural point of view, the layer is not limited to typical DNN layers, such as convolution or fully-connected layers, but can take any form for transformation, which is trainable by e.g. backpropagation.

In our case, we opted for linear transformations due to their simplicity.
One interesting property of linear transformations in this use case is, that they can be merged with adjacent linear transformations without any difference in the overall result.
Since DNNs commonly consist of convolution and fully-connected layers, the integration of the transformation into existing weights is possible.
In terms of computational complexity, the DNN inference generalized with merged CL is identical to its non-generalized baseline.

In our study, we implemented two types of CL: linear channel-wise and the linear inter-channel transform.
\begin{align}
    \mathbf{f}_\text{cw}(\mathbf{x}) &= (\mathbf{w} + \boldsymbol{1}) \odot \mathbf{x}\label{eq:cw}\\
    \mathbf{f}_\text{ic}(\mathbf{x}) &= (\mathbf{W} + \mathbb{I}) \cdot \mathbf{x}\label{eq:ic}
\end{align}
As noted in Eq. \eqref{eq:cw} and \eqref{eq:ic}, CW scales each channel seperately and IC additionally includes inter-channel dependencies.
Here, IC offers more degrees of freedom to correct joint distribution discrepancies than CW, but also uses more trainable parameters.



\subsection{CL QoS Robustness}
\label{ssec:cl_qos_robust}

To evaluate proposed CL, we chose the CNN architecture from \cite{Loh2022}, as it can solve the full CINC'17 benchmark with state-of-the-art QoS.
Then, we trained the architecture for the binary classification use case with the AFDB data.
Here, the data is split into SD and TD, such that a pre-trained model from the SD can be evaluated and then fine-tuned on the TD.
The subsets are chosen by combining data of each patient, such that the TD contains balanced number of labels\footnote{The number of labels from each class differs by max. \SI{5}{\percent}}, while the SD consists of the remaining patients.
The balanced TD is necessary to ensure good fine-tuning performance on small sample sizes. 
The experiment is repeated for all permutations of one or two patients that guarantee balanced TD to generate statistically robust results.
The evaluation within the TD subset is performed using stratified 5-fold cross-validation to ensure input data independent CL training performance.
Note that in this experiment we specifically focus on inter-patient domain shift without loss in generality.

An important item to consider is that our experiment assumes the availability of TD data for training to explore the maximum capacity of CL in the context of DG.
The modification of CL training towards unsupervised methods is expected to yield reduced QoS, but can be performed using metrics, which quantify the quality of generalized features and can be used as the loss function during training \cite{Wang2021}.
However, this is not investigated within the scope of this work.


\subsection{Results and Discussion}
\label{ssec:cl_results}

\begin{figure}[ht]
    \centering
  \subfloat[Insertion of a single channel-wise CL after each layer\label{fig:cl_qos_cw}]{%
      \includegraphics[width=0.485\textwidth]{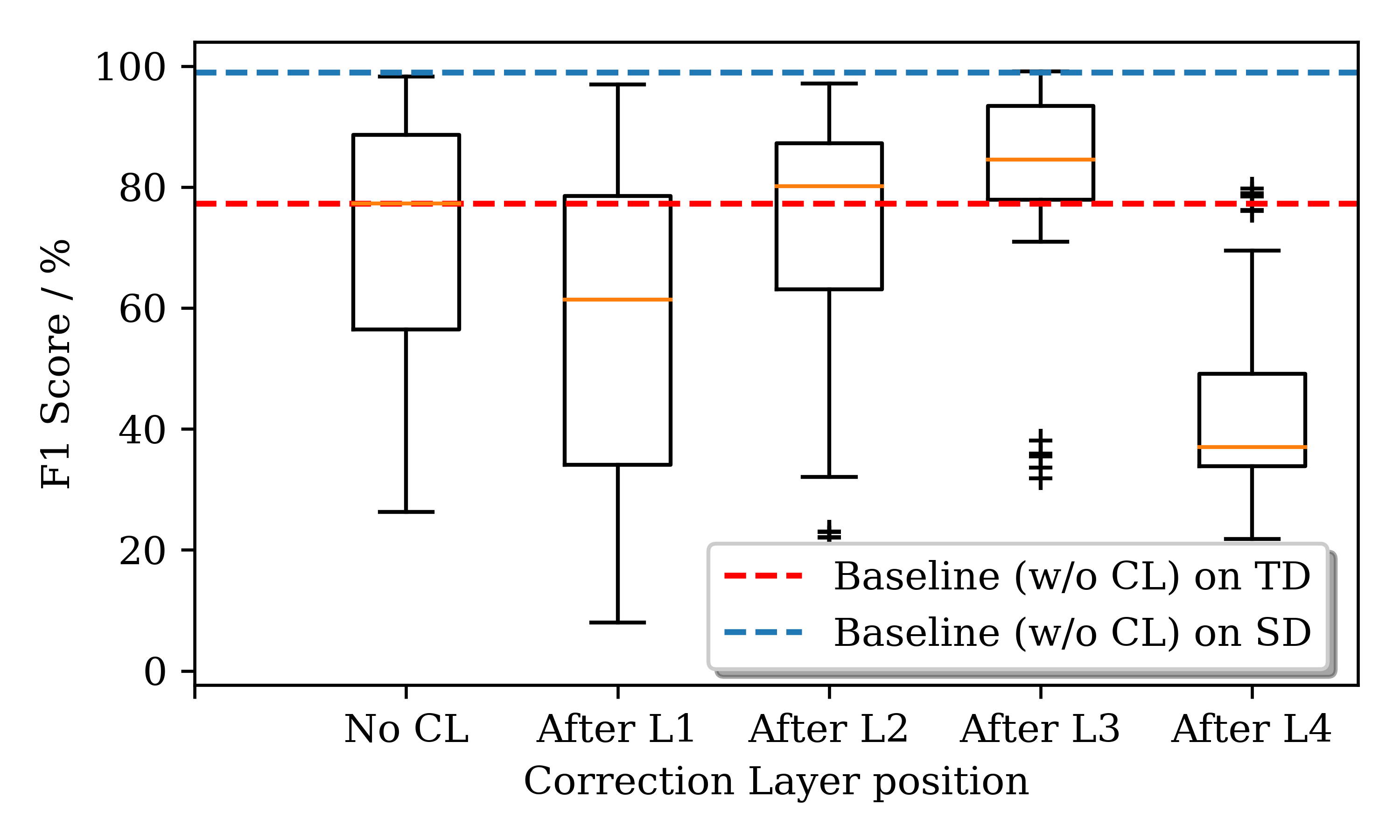}}
    \hfill
  \subfloat[Insertion of a single inter-channel CL after each layer\label{fig:cl_qos_ic}]{%
        \includegraphics[width=0.485\textwidth]{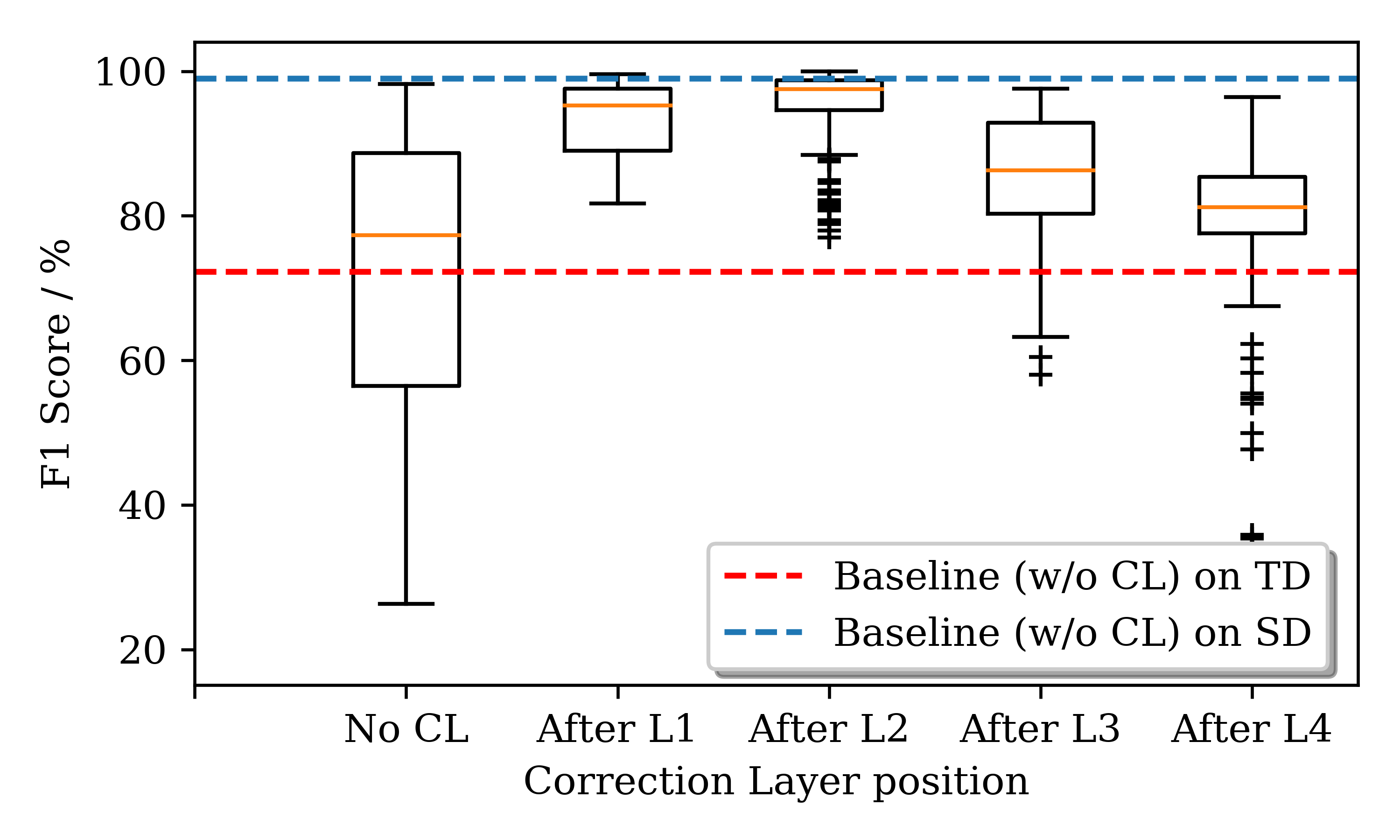}}
    \\
  \caption{Evaluation of CL performance regarding the type and position using CNN architecture from \cite{Loh2022}. The blue and red line indicate the average performance on the SD and TD w/o CL training.}
  \label{fig:cl_qos} 
\end{figure}

The performance on SD, TD before as well as TD after CL training is illustrated in Fig. \ref{fig:cl_qos} for all positions of the CL in between existing DNN layers.
It is evident that all models perform very well on the SD (average: blue line) but the performance drops significantly on the unseen data from the TD (average: red line).
Looking at the channel-wise CL, the improvement on QoS, i.e. F1-score, is selective to specific positions of the CL with some even underperforming the non-modified baseline.
The inter-channel CL, however, shows consistent improvements in the F1-score, especially, when inserted in the center position of the baseline CNN with $\Delta F1 = 21.16\,\%$ compared to the average.
Even though the choice of used dataset is critical for a general interpretation (as discussed in Section \ref{ssec:ecg_dg}), the trend is robust for all combinations across train and test set patients in the investigated setup.

We also observe, that a larger degree of freedom, i.e. more trainable parameters in the CL, yield better results, since inter-channel CL clearly outperforms channel-wise CL.
The good performance in the center position of Fig.\ \ref{fig:cl_qos_ic} reflects previous observations about two opposing trends in feature distributions for DNN \cite{Yosinski2014}.
On the one hand, the task specificity of intermediate activations increase for layers close to the output, thus, decreasing the generalizing capability of those features.
On the other hand, the discrepancies between input data decrease the deeper the intermediate features are relative to the input.
In principle, the maximum capacity for generalization is, consequently, in the central layers of the DNN.

Although the results in Fig.\ \ref{fig:cl_qos} are based on CL training on the whole TD, we investigated whether the training process can be further simplified.
To observe the necessary training samples for generalization, we repeated the experiment with less training samples.
Specifically, we reduced the number of training samples per recording without any issues in training convergence, whilst still keeping the same experiment setup as described in Section \ref{ssec:cl_qos_robust}.
In the first try, a $2.99\,\times$ reduction of training samples can be achieved with only 1\,\% F1 score reduction.
In the extreme case, even a $120.48\,\times$ smaller training dataset is possible while tolerating an acceptable reduction of 6\,\%.
This result shows, that CL training does not require many training samples to yield acceptable QoS performance.
Hence, test-time training of CL is a competitive option compared against state-of-the-art ECG DG methods utilizing active datasets, e.g. \cite{Rahhal2016,Wang2019}, whilst still changing only CL parameter.
Another big feature is that the chosen transformation is linear and can be merged with convolution and fully-connected layers (further evaluated in Section \ref{ssec:cl_hw_forward}).
It means that training is simply reduced to a single layer only, while it is observed in earlier experiment that this single layer fine-tuning requires only few samples.
Consequently, the proposed CL showcases that constrained DNN training both in terms of model parameters and input samples yield state-of-the-art DG performance in the addressed usecase.
In the end, we have successfully proven that the emerging class of HW-optimized DG algorithms has the potential to address the problem of DG with nearly negligible classification performance degradation compared to conventional approaches.

\section{Hardware Complexity}
\label{sec:cl_hw}

In a next step, we validate the predicted hardware efficiency of CL.
As a reference for comparison, the status-quo of both DNN inference and training is considered.
In the former, the integration of CL is investigated in an ultra-low power ECG DNN inference engine \cite{Loh2022}.
In the latter, on-device training is evaluated using high-level estimations for CC and memory overhead.


\subsection{Integration of Correction Layer in an ECG accelerator}
\label{ssec:cl_hw_forward}

In the following, we implemented two logic designs for comparison: A reference design without CL (Ref) and a design with CL as a dedicated layer.
The designs are synthesized in a \SI{22}{\nano\meter} CMOS technology using commercial EDA tools, i.e. Synopsys Design Compiler.
Both post-synthesis netlists are simulated at \SI{0.8}{\volt} (nominal) and typical process corners at \SI{2}{\mega\hertz}.
To estimate the energy required to process each input sample the traces are recorded in the active calculation period.
In our experiment, we chose the period of highest computational load, i.e. input sample triggering computation throughout the entire DNN until prediction sample is generated.
Note that this worst case scenario deviates from the typical case as evaluated in \cite{Loh2022}, since it does not consider the idle periods in between input samples.
This test configuration was selected to derive relevant quantitative comparison results while requiring reasonable simulation time and complexity.

The baseline HW architecture consists of activation memory, weight memory, a vector PE and control/multiplexing logic for activation routing.
In the reference architecture, the activation memory and weight memory is split into shift registers and SRAM respectively, due to stream processing architecture and the utilized output stationary dataflow to reuse the vector PE element.

The integration of an additional layer, therefore requires simply a larger SRAM for the additional weights and another set of shift registers for the input activations of the CL.
A key difference is the mapping of the matrix-vector multiplication (from Eq. \eqref{eq:ic}) to the existing vector PE.
This is achieved through loop tiling and corresponding control logic modifications.
For instance, the computation of one row of a $24\times 24$ matrix with the input vector of 24 elements is treated the same as a $5\times 1$ convolution with $5$ input channels.
The number of output elements, i.e. 24 rows of the weight matrix, correspond to the number of output channels in the convolution and the output samples are fed into the shift registers of the next layer.

Similar to the batch normalization layers, adjacent linear operations, such as the matrix-vector multiplication of the CL and the convolution of a convolution layer, can be merged into a single operation, while still computing the arithmetically identical output activations.
Conceptually, this results in a third design, i.e. the CL merged into the adjacent convolution layer.

\begin{table}[htb]
	\centering
	\caption{Post-synthesis results of reference ECG accelerator with and without integrated CL}
	\label{tab:dg_synth_comp}
	\centering
    \begin{tabular}{|C{0.069\textwidth}||C{0.03\textwidth}|C{0.03\textwidth}||C{0.03\textwidth}|C{0.03\textwidth}|C{0.03\textwidth}||C{0.033\textwidth}||C{0.033\textwidth}|}
    \hline
    & Train Acc. (\SI{}{\percent}) & Test Acc. (\SI{}{\percent}) & Area (\SI{}{\micro\meter^2}) & Seq. Cells & Comb. Cells & Max. Freq. (\SI{}{\mega\hertz}) & Energy (\SI{}{\nano\joule}) \\ \hline\hline
    CL Design & \multirow{ 2}{*}{98.98} & \multirow{ 2}{*}{97.16} & 19.9k & 19.7k & 6.9k & 9.96 & 68.21 \\ \cline{1-1}\cline{4-8} 
    CL Design (integr.)&  &  & \multirow{ 2}{*}{19.3k} & \multirow{ 2}{*}{19.4k} & \multirow{ 2}{*}{6.6k} & \multirow{ 2}{*}{9.97} & \multirow{ 2}{*}{65.14} \\ \cline{1-3} 
    Reference Design & 99.01 & 95.15 &  &  &  &  &  \\ \hline
    \end{tabular}
\end{table}

Table \ref{tab:dg_synth_comp} shows the post-synthesis results of the implemented logic designs.
It is evident, that the CL design performs more robust across both train and test set generalizing across patient groups.
If the CL is implemented explicitly, more sequential cells are synthesized to store CL input activations and more combinational logic is necessary for control/routing logic.
Therefore, the overall area and energy is increased slightly by \SI{3.1}{\percent} and \SI{4.7}{\percent}, respectively.
In the case that CL is merged into the convolution, the reference design can be reused with different weights in one layer.
In terms of the design's HW key performance indicators (KPIs), they are identical to the reference design with the classification performance of the CL design.
Hence, no compromises need to be made for CL integration in the DNN inference.

\subsection{Backpropagation On-Chip}

Considering inference alone does not reveal the strength of proposed CL.
Therefore, in a next step, we estimate expected CC and memory requirements of CL in comparison to conventional fine-tuning during deployment.

As detailed in \cite{Lee2021}, the weight updates in the training step are calculated based on the input activations $x^{(l)}_i$, weights $w^{(l)}_{ij}$ and the gradients of each layer $l$, where $i,j$ indicate the neuron in the input and output of the layer, respectively.
In specific, the weight increment $\Delta w^{(l)}_{ij}$ is calculated using one variant of the stochastic gradient descent algorithm, which depend on $x^{(l)}_i$ and $w^{(l)}_{ij}$ through the gradient of the loss function $\partial L/\partial w^{(l)}_{ij}$ (see Eq. \eqref{eq:w_update} and \eqref{eq:grad}).

\begin{align}
    w^{(l)}_{ij} &\rightarrow w^{(l)}_{ij} + \Delta w^{(l)}_{ij} \label{eq:w_update}\\ 
    \frac{\partial L}{\partial w^{(l)}_{ij}} &= \frac{\partial L}{\partial x^{(l)}_{j}}\frac{\partial x^{(l)}_j}{\partial w^{(l)}_{ij}} \label{eq:grad}
\end{align}

\begin{figure}[ht]
    \centering
    \includegraphics[width=0.485\textwidth]{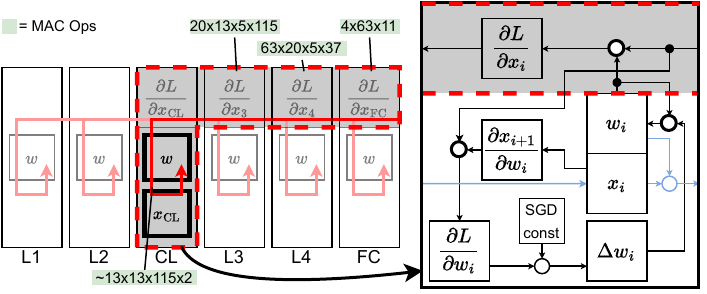}
    \caption{Simplified concept of CL CC and memory requirements in the example CNN from \cite{Loh2022}. Additional memory is marked in bold solid lines and the CC for training is indicated in the area marked with red dashed lines. The MAC operations for each subcomponent is highlighted in green.}
    \label{fig:bp_concept_cc_mem}
\end{figure}

Figure \ref{fig:bp_concept_cc_mem} visualizes the memory and CC reductions of CL training compared to conventional DNN fine-tuning reference in an example.
For the reference, all partial derivatives need to be calculated in all layers and, therefore, all activations need to be stored as well for the weight update.
For CL training, only the weights and activations of the CL need to be stored in addition to the reference DNN, while the partial derivatives $\partial L/\partial x^{(l)}_{j}$ required for calculation do not need permanent buffers.
They are calculated recursively starting from the loss at the output layer, while the activations of other layers are not required for their computation.
We approximate CC with multiply-and-accumulate (MAC) operations similar to complexity estimations in the forward pass \cite{Parashar2019}.
We further consider different layer types, such as fully connected (FC) layers and convolution (CONV) layers with and without subsequent pooling, to increase the granularity of our estimation model.

\begin{figure}[ht]
    \centering
  \subfloat[Normalized MAC operations for training over the position of inserted CL in the DNN (left: input, right: output)\label{fig:hw_eval_mac}]{%
      \includegraphics[width=0.75\linewidth]{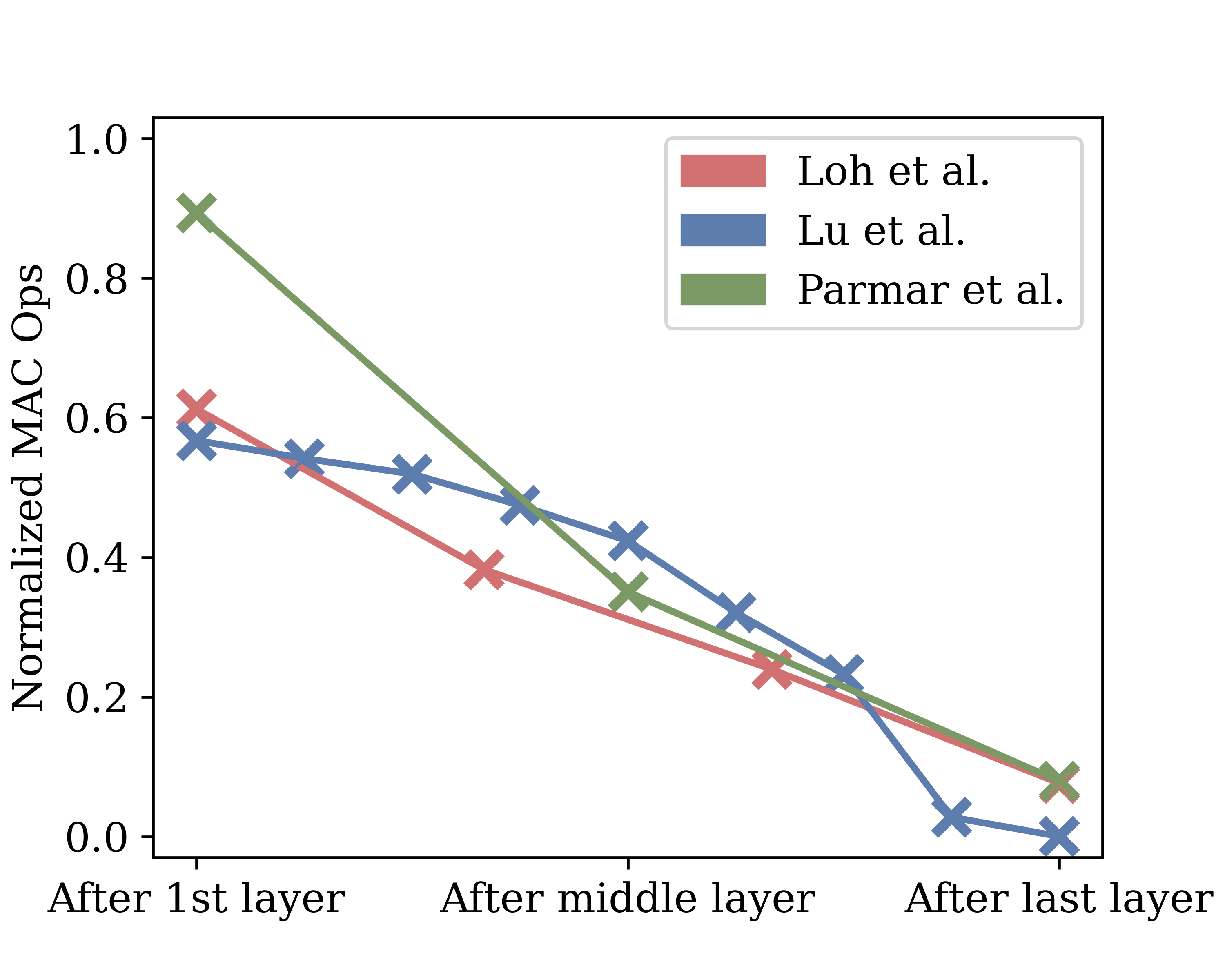}}
    \hfill
  \subfloat[Normalized memory overhead for training over the position of inserted CL in the DNN (left: input, right: output)\label{fig:hw_eval_mem}]{%
        \includegraphics[width=0.8\linewidth]{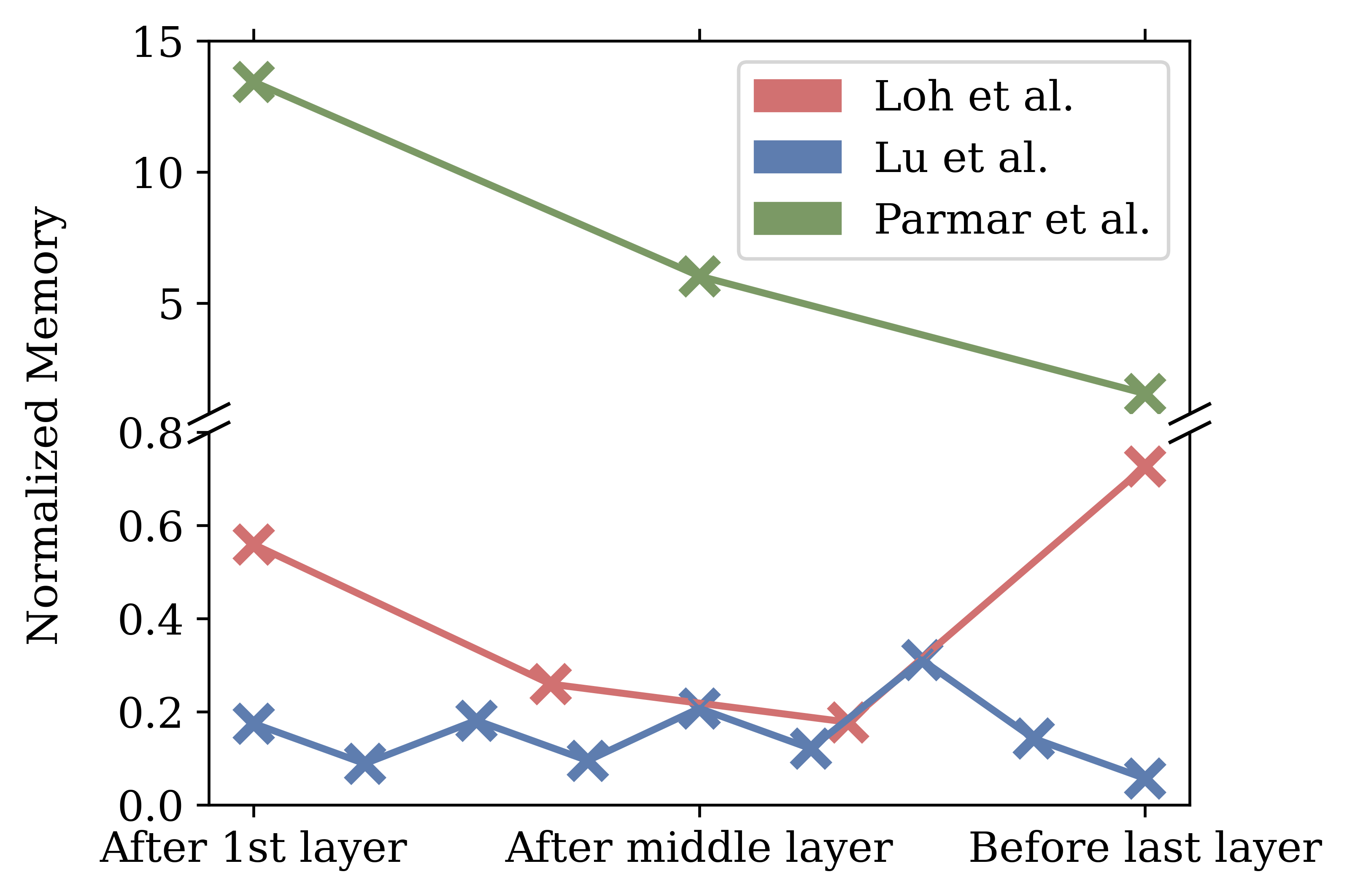}}
    \\
  \caption{Memory and CC of backpropagation for HW accelerators Loh et al. \cite{Loh2022}, Lu et al. \cite{Lu2021} and Parmar et al. \cite{Parmar2023}. The complexity is normalized against the reference case, in which the whole DNN is fine-tuned. }
  \vspace{-0.4cm}
  \label{fig:hw_eval} 
\end{figure}

Figure \ref{fig:hw_eval} shows results of CC and memory estimations for different positions of the CL within a variety of DNN architectures, which are suitable for HW acceleration on embedded platforms.
In this comparison, inter-channel CL are used as an example, since they showed the best QoS performance in the evaluation in Section \ref{sec:cl_alg}.

In general, fine-tuning the complete model requires most MAC operations compared to the insertion of a CL regardless of the DNN architecture (see Fig. \ref{fig:hw_eval_mac}).
Further it is visible, that the number of MAC operations decrease the closer the CL is inserted to the final output layer.
This trend is a consequence from the recursive calculation of $\partial L/\partial x^{(k)}_{j}$.
The overall CC, however, is dependent on the DNN model architecture, as CL size is dependent on the dimensions of its input.

The additional memory required for CL, however, does not show such consistent trend (see Fig. \ref{fig:hw_eval_mem}).
While CL in the CNN architectures from \cite{Lu2021} and \cite{Loh2022} show consistently less memory overhead than the DNN fine-tuning, the MLP architecture of \cite{Parmar2023} includes a higher memory requirement for CL training.
Here, the number of parameters necessary for the CL, i.e. square of intermediate activitions of that layer, exceed the storage saved by omitting the buffer for the frozen layers.
As the relative difference of CL parameters decrease for deeper networks, this trend is an exception for shallow DNNs.
Furthermore, the memory overhead is mainly dependent on the size of inserted CL, i.e. its input dimensions.
For instance, the intermediate activations in the CNN model of \cite{Loh2022} consistently decrease the closer the CL is inserted to the output layer.
However, the memory increases when inserted directly before the final FC layer, as its input dimensions, i.e. feature dimension and input channels, are much larger than in the previous layers.
A different example is the model of \cite{Lu2021}, where the input dimensions of all layers are quite consistent.
Here, the aforementioned memory reductions with max pooling layers are visible in nearly every second layer.

All in all, a trade-off needs to be found between achieved QoS, CC and memory.
Considering the best design point from the QoS evaluation (see Section \ref{sec:cl_alg}), i.e. CL after L2, a CC reduction of $2.5\,\times$  and a memory reduction of more than $3\,\times$ is observed.
Hence, CL training provide a more efficient alternative to fine-tuning, while achieving competitive classification performance on the target domain.

	\section{Conclusion}
\label{sec:conclusion} 


Practical deployment of DNN models in mobile devices require robust methods to cope with the domain shift problem.
The main challenge is the combination of the two independent disciplines of DNN HW acceleration and DG approaches on the algorithm level.
Especially, sensitive applications such as ECG classifications for health monitoring benefit from algorithm-hardware co-designed systems for DG on-the-edge, as domain shift is inherent in practical scenarios and privacy concerns limit the data transmission to central servers.
To address this challenge, this work presented DG methods specifically for the deployment on HW accelerators using ECG processing as the application context.
The variety of DG methods require modifications on the pre-trained DNN, while the majority aims to fine-tune the pre-trained DNN to generalize across domains.
In the case of ECG classification, the scarcity of data and high quality labels limit the capability of DG methods to fully capture all variants of domain shift, although state-of-the-art methods are capable of solving inter-patient domain shift effectively.
As a first step towards co-optimized DG methods, we introduced ``correction layers'' (CLs) as a low complexity DG method to solve inter-patient domain shift.
This is achieved by freezing a pre-trained DNN, while adding a single trainable CL for feature normalization.
Our evaluation shows that CL with inter-channel transformations provide robust QoS improvements of $\Delta F1 \ge 20\,\%$.
These improvements in classification performance require no hardware overhead during inference and are complemented by CC and memory reductions of more than $2.5\,\times$ and $3\,\times$ during training, respectively.
In the end, the CL study shows that the co-optimization of algorithm and hardware yield state-of-the-art ECG classification results on inter-patient DG with minimal modification on existing ECG accelerators.

	\bibliographystyle{IEEEtran}
	\bibliography{IEEEabrv,latex/bstcontrol,latex/user_bib}

\end{document}